\documentclass[letterpaper, 10 pt, conference]{ieeeconf}  

\IEEEoverridecommandlockouts                              

\usepackage{graphics} 
\usepackage{epsfig} 
\usepackage{mathptmx} 
\usepackage{times} 
\usepackage{amsmath} 
\usepackage{amssymb}  
\usepackage{lineno}
\usepackage{amsfonts}
\usepackage{multirow}
\usepackage{array}
\usepackage{graphicx} 
\usepackage{url}

\title{\LARGE \bf
SlipNet: Enhancing Slip Cost Mapping for Autonomous Navigation on Heterogeneous and Deformable Terrains}

\author{Mubarak Yakubu$^{1,2}$, Yahya Zweiri$^{1,2}$, Ahmad Abubakar$^{1,3}$, Rana Azzam $^{1,3}$, Muhammad Humais$^{1,2}$,\\ Ruqayya Alhammadi $^{1,3}$, and Lakmal Seneviratne $^{1,3}$ 
\thanks{*This research publication was funded by the Khalifa University Center for Autonomous Robotic Systems under Award No. RC1-2018-KUCARS}
\thanks{$^{1}$  M. Yakubu, Y. Zweiri, A. Abubakar, R. Azzam, R. Alhammadi, and L. Seneviratne are with the Khalifa University Center for Autonomous and Robotic Systems (KUCARS), (100060013, yahya.zweiri, 100059792, 100041348, lakmal.seneviratne)@ku.ac.ae. $^{2}$ Y. Zweiri is also associated with the Advanced Research and Innovation Center (ARIC) and the Department of Aerospace Engineering, Khalifa University. $^{3}$ L. Seneviratne is also with the Department of Mechanical Engineering at Khalifa University, Abu Dhabi, United Arab Emirates (Corresponding author's email: 100059792@ku.ac.ae)}
}

\begin{document}

\maketitle
\thispagestyle{empty}
\pagestyle{empty}

\begin{abstract}

Autonomous space rovers face significant challenges when navigating deformable and heterogeneous terrains due to variability in soil properties, which can lead to severe wheel slip, compromising navigation efficiency and increasing the risk of entrapment. To address this problem, we introduce SlipNet, a novel approach for predicting wheel slip in segmented regions of diverse terrain surfaces without relying on prior terrain classification. SlipNet employs dynamic terrain segmentation and slip assignment techniques on previously unseen data, enhancing rover navigation capabilities in uncertain environments. We developed a synthetic data generation framework using the high-fidelity Vortex Studio simulator to create realistic datasets that replicate a wide range of deformable terrain conditions for training and evaluation. Extensive simulation results demonstrate that our model, combining DeepLab v3+ with SlipNet, significantly outperforms the state-of-the-art TerrainNet method, achieving lower mean absolute error (MAE) across five distinct terrain samples. These findings highlight the effectiveness of SlipNet in improving rover navigation in challenging terrains.

\end{abstract}

\section{INTRODUCTION}

Autonomous navigation is crucial for space exploration missions, where robots operate independently on planetary surfaces with minimal or no human intervention. The significant communication delays between Earth and distant planets, such as Mars, necessitate reliable onboard sensors that can perceive the surrounding environment, assess traversability, and navigate safely~\cite{quadrelli2015guidance}. Previous Mars missions have equipped rovers with stereo-vision systems to avoid obstacles and plan paths autonomously. However, these rovers often require human intervention to assess traversability, resulting in slow progress. For instance, the Curiosity rover's drive distance was limited due to significant slippage on deformable surfaces, caused by the looseness of the sand, leading to significant path deviation \cite{toupet2020driving}. These surfaces, which can vary greatly in soil composition, are examples of heterogeneous terrains. Such terrains exhibit variations in soil texture and physical properties, making navigation even more challenging. Overcoming these challenges is crucial for enhancing the efficiency and safety of rovers during space exploration missions.


Traversability assessment on deformable terrains, particularly for space environments, is challenging due to the complex dependencies between physical terrain properties, surface geometry, and the rover's mobility mechanisms. These dependencies vary widely across different terrains, making it difficult to generalize from one terrain type to another. Existing methods often rely on visual sensors to classify terrain types, but these methods can fail when encountering new, unseen terrains, leading to misclassification and prediction errors \cite{arvidson2017mars, gonzalez2018slippage}. Factors like sand density, rock friction, and terrain unevenness further complicate traversability assessment, influencing wheel slip and vehicle dynamics and necessitating reliable slip prediction models \cite{hutangkabodee2006performance}. This complexity is further exacerbated in extraterrestrial environments, where the lack of prior terrain knowledge can hinder real-time decision-making. For example, reliance on visual sensors alone might be problematic, as visual cues may not fully capture the physical properties affecting slip, such as soil moisture or compaction \cite{10406669}. These properties influence the interaction between the rover's wheels and the terrain, affecting traction and potentially leading to slip \cite{song2009slip}.

Machine learning methods, particularly deep learning, have been used to tackle these challenges. Convolutional Neural Networks (CNNs) have demonstrated effectiveness in feature extraction and terrain classification, leveraging large datasets to learn unique patterns from terrain images \cite{guan2022ga,guan2021tns}. However, these methods often require substantial amounts of annotated data, which can be challenging to obtain for planetary surfaces \cite{maturana2018real}. Also, they could not capture the soil's physical parameters. Hence, they fail to generalize well to unknown terrains due to the variability in soil properties. Self-supervised learning approaches have emerged as a solution, associating sensor data with terrain properties to enable models to learn from real-world interactions without extensive annotations~\cite{wellhausen2019should}. Despite these advancements, accurately predicting wheel slip remains difficult due to the complex dynamic nature of terrain interactions. Factors such as the interaction between the wheel and the soil, which can vary significantly, highlight the need for more robust and adaptable methods \cite{rothrock2016spoc}.
\begin{figure}
    \centering
    \includegraphics[width=9.0cm]{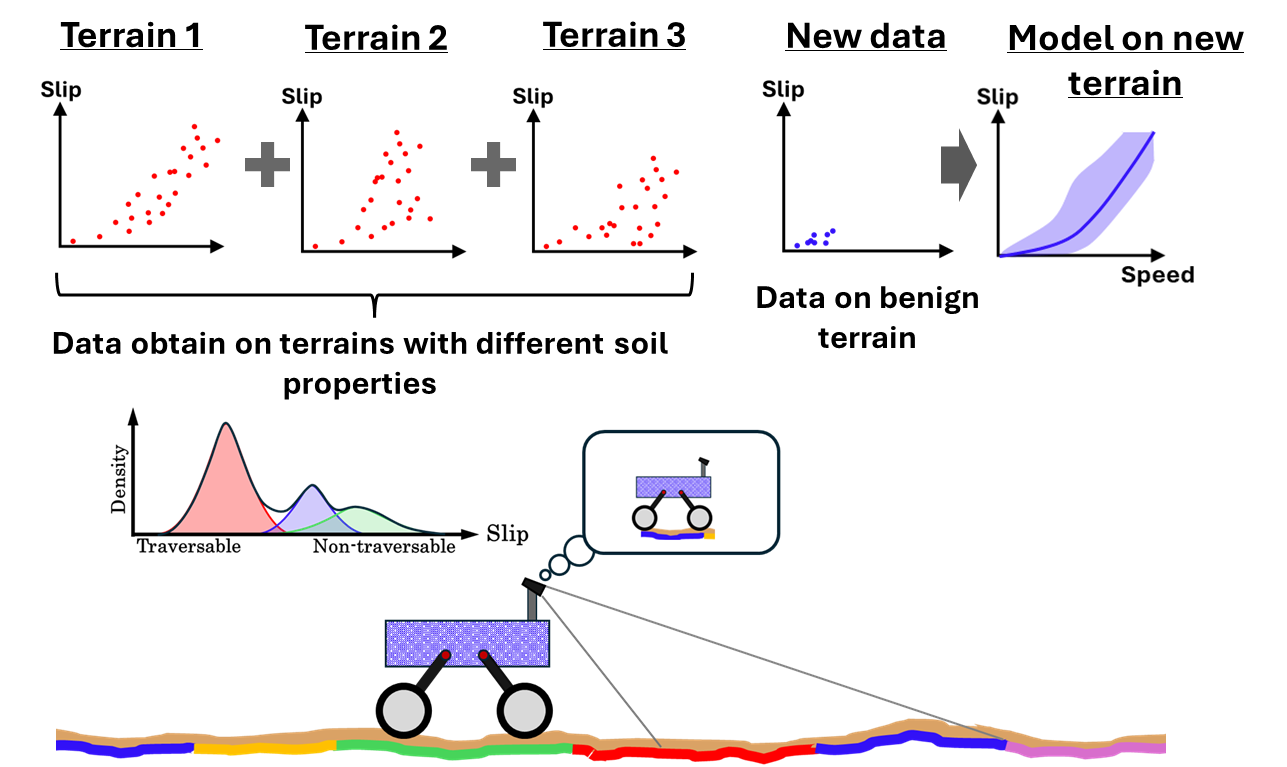}
    \caption{Concept of the proposed SlipNet that takes as input multiple terrain semantic images and in-situ wheel slip and speed measurements. By utilizing past experiences on various terrain surfaces, the proposed method enhances prediction accuracy on potentially hazardous terrains in new environments, where training data is limited or not available.}
    \label{fig:proposed_concept}
\end{figure}
 
This paper proposes SlipNet, an approach for real-time terrain classification and slip prediction applicable to autonomous navigation on deformable terrains. SlipNet dynamically reclassifies unseen terrains during deployment, enabling the rover to adapt to new and evolving terrain conditions (Fig. \ref{fig:proposed_concept}). 
Our method uses a dual-network architecture, with the segmentation network providing terrain classification and the Slip Risk module contributing slip prediction. The system integrates sensor data and camera input to generate a slip cost map (SCM) that aids in navigation.
The main contributions of the paper are summarized as follows:

\begin{enumerate}

\item We introduce SlipNet, a novel dual-network architecture that fuses real-time terrain segmentation with probabilistic slip prediction. By integrating visual perception with terrain slippage characteristics, SlipNet enables robust and accurate autonomous navigation over deformable terrains.

\item We develop a synthetic data generation framework leveraging the high-fidelity Vortex Studio simulator. This framework produces realistic datasets that replicate a wide range of deformable terrain environments, facilitating the training and evaluation of SlipNet under diverse conditions.

\item  We conduct extensive simulation experiments on unseen terrains with varying soil properties, demonstrating that SlipNet significantly outperforms the state-of-the-art method, TerrainNet~\cite{meng2023terrainnet}, by achieving lower mean absolute error (MAE) across all test scenarios.

\end{enumerate}
The paper's structure is as follows. \textbf{Section I} delves into the concept of terrain traversibility assessment. \textbf{Section II} presents our methodology in slip risk levels categorization, data generation, and model architecture. \textbf{Section III} discusses a summary of the simulation results. Finally, \textbf{Section IV} provides a conclusion and outlines potential directions for future research.

\subsection{Related Work}
\label{sec:related_work}
In this section, we present a survey of current practices in slip prediction and traversability estimation based on semantic segmentation and self-supervised techniques. 


\subsubsection{Current practice on Mars}
Currently, slip prediction for daily tactical planning is a manual process on Earth. Rover operators begin by visually identifying the terrain type, then estimating slip based on slope versus slip curves specific to each terrain class, such as loose sand, consolidated sand, and bedrock, which have been derived from comprehensive earth-based testing \cite{heverly2013traverse}. While these models are generally accurate for most terrains, they often underestimate the variability in slip. Moreover, the models do not specifically address longitudinal or lateral slips; instead, they assume the slip vector always points down-slope, which can lead to a significant overestimation of lateral slip. Efforts are underway to develop automated systems for visual terrain classification and slip estimation using data collected directly from rovers \cite{rothrock2016spoc}.

For autonomous navigation, rovers employ the Grid-based Estimation of Surface Traversability Applied to Local Terrain (GESTALT) system \cite{goldberg2002stereo}, which protects against geometric obstacles but does not predict slip. During autonomous operation, rovers estimate slip by comparing the distance covered as measured by visual odometry (VO) against the expected distance without a slip from wheel odometry. If the detected slip exceeds a certain threshold, the rover halts and awaits further instructions~\cite{maimone2007two}. The onboard imaging systems primarily focus on identifying targets for scientific instruments, as seen in the Autonomous Exploration for Gathering Increased Science (AEGIS) system on the Opportunity rover, which identifies scientific targets either by recognizing rocks with defined edges \cite{estlin2012aegis} or using a random forest classifier \cite{wagstaff2013smart}. However, these systems do not evaluate the traversability classes.


\begin{figure*}
    \centering
    \includegraphics[width=1\linewidth]{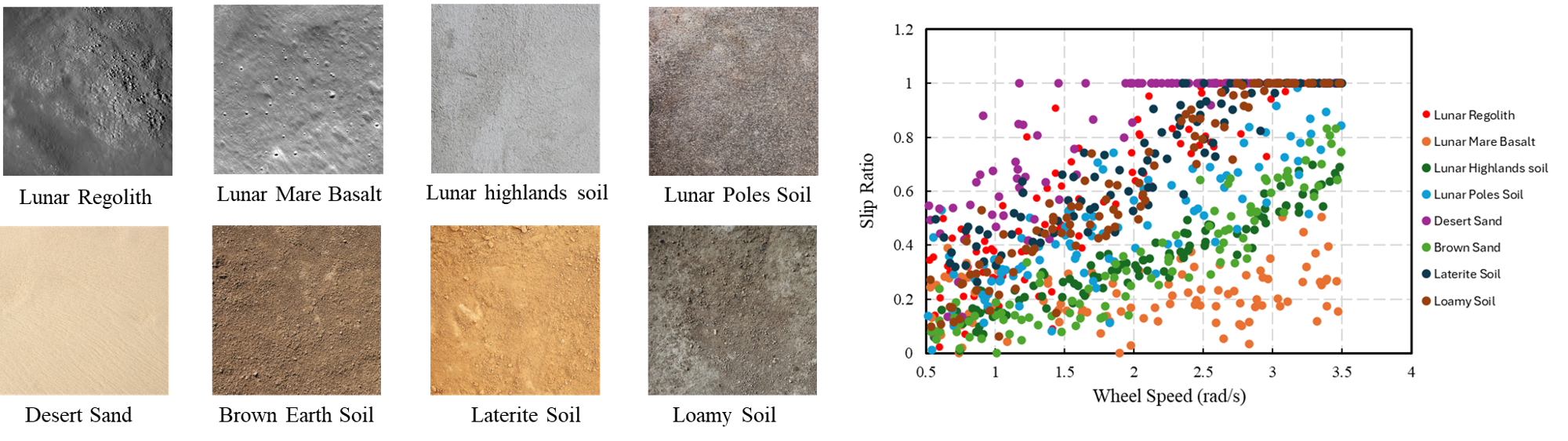}
    \caption{Slip versus Speed dataset generated for eight different terrain types}
    \label{fig:slip-speed_data}
\end{figure*}

\subsubsection{Traversability from Semantics}
Semantic segmentation is essential in computer vision, involving pixel-wise image labeling, and has been enhanced by deep learning architectures like OCRNet \cite{gupta2021ocrnet} and PSPNet \cite{zhao2017pyramid}. Recent transformer-based methods offer superior performance but require significant computational resources, prompting the development of more efficient designs \cite{dai2022segmarsvit}. These methods excel in structured datasets like CityScapes but struggle with deformable terrain environments due to indistinct boundaries and class features.

Deep learning excels in semantic scene understanding from both images \cite{viswanath2021offseg,roth2023viplanner}, and point clouds \cite{shaban2022semantic}. Semantic classes are mapped and linked to traversability scores using datasets like RUGD \cite{wigness2019rugd}, Rellis-3D \cite{jiang2021rellis}, and Freiburg Forest \cite{valada2017deep}, though their limited size and diversity constrain broad applicability and accuracy. Techniques range from voxel-wise terrain density classification using LiDAR and cameras \cite{bradley2015scene} to projecting image-based semantic segmentation onto 2.5D maps \cite{maturana2018real}, and combining LiDAR with learned semantics for region risk assessment \cite{schilling2017geometric}. Space exploration faces similar challenges in terrain hazard identification, critical for rover missions \cite{ono2015risk}. Efforts to classify terrain types and their physical interactions include using CNNs to predict wheel slip on Mars \cite{rothrock2016spoc} and developing comprehensive datasets \cite{swan2021ai4mars}. Research also focuses on reducing training data needs \cite{zhang2022s} and probabilistically merging semantic classifications with slip models for risk-aware traversability \cite{endo2023risk}.

\subsubsection{Self-supervision} Traversability estimation methods that depend on scene semantics often require costly annotated data. Self-supervised approaches address this challenge by generating training signals without manual annotations, utilizing information from alternative sensor modalities \cite{meng2023terrainnet} or from robot-environment interactions \cite{cai2023probabilistic}. These self-generated supervision signals enable models to predict future terrain conditions without direct interaction with the terrain. In  \cite{otsu2016autonomous}, two classifiers were enhanced for terrain class prediction using image and proprioceptive data in a space exploration context. The work in \cite{castro2023does} adjusted proprioceptive-based pseudo labels using vibration data to predict traversability from colorized elevation maps. During indoor navigation, \cite{richter2017safe} suggested using anomaly detection to identify safe image areas. Other researchers have applied anomaly detection and evidential deep learning to learn from real-world data without manual labels \cite{frey2023fast,schmid2022self}. Contrastive learning has also proved effective in creating expressive representations for traversability estimation \cite{xue2023contrastive} generated heuristic-based and velocity-tracking pseudo labels, respectively, enabling online training of traversability prediction networks during deployment. \cite{gasparino2022wayfast} in WayFAST approximated terrain traversability using tracking errors from a predictive control model, and \cite{cai2023probabilistic} estimated worst-case costs and traction from semantic elevation maps, including confidence values via density estimation.

\section{Methodology}

\subsection{Background on Slip}
\label{sec:slip_calculation}
This work examines the longitudinal slip (along the travel direction) of the rover as in \cite{10406669,alhammadi2024event}. The longitudinal slip is defined as:

\begin{equation}
s = \begin{cases} 
\frac{v_x - v_{\text{ref}}}{v_{\text{ref}}}, & \text{if } v_x < v_{\text{ref}} \text{ (driving)} \\
\frac{v_x - v_{\text{ref}}}{v_{\text{ref}}}, & \text{if } v_x > v_{\text{ref}} \text{ (braking)}
\end{cases}
\end{equation}
\noindent where \(v_x\) is the rover's measured velocity in the direction of travel and \(v_{\text{ref}}\) is the commanded velocity. A positive slip means the rover is traveling slower than commanded, and a negative slip means the rover is traveling faster than commanded. This slip value, $s$ ranges between -1 and 1. In this paper, only the positive slip values are considered. 

\subsection{Slip Risk Module}
As mentioned in Section \ref{sec:slip_calculation}, the slip ratio quantitatively describes the wheel slippage. The slip ratio can be classified into five different categories: ($0 < s_1 \leq 0.2$), ($0.2 < s_2 \leq 0.4$), ($0.4 < s_3 \leq 0.6$), ($0.6 < s_4 \leq 0.8$), and ($0.8 < s_5$). The simulation results of the current study, as presented later, indicate that at low and moderate wheel speeds, the rover experiences slip values up to 0.4, but encounters high and unpredictable slip as the wheel speed increases significantly. Slip measures the navigation risk encountered by wheeled mobile robots on deformable terrains, which has been studied in previous literature \cite{10634320}. Before predicting risk, the slip ratio is estimated from a regression curve using collected slip versus speed data. The estimated slip ratio, denoted as $s$, and a linear basis function $y(x,w)$ as in (\ref{eq:slip_regression}).

\begin{equation}
\label{eq:slip_regression}
    s = y(\textbf{x, w}) + \epsilon,
\end{equation}

\begin{equation}
    y(\textbf{x, w}) = \sum_{j=0}^{M-1} w_j \phi_j(\textbf{x}) = w^T \phi(\textbf{x}).
\end{equation}
where $\epsilon$ is the prediction error, $\textbf{x}$ is the wheel speed variable, $\textbf{w}$ is the weight vector and $\phi(\textbf{x})$ are the basis functions. To minimize error, $\textbf{w}$ is defined in (\ref{eq:min_error}).

\begin{equation}
 \label{eq:min_error}
    \min E(w) = \frac{1}{2} \sum_{n=1}^N \left\{ s - w^T \phi(\textbf{x}) \right\}^2.
\end{equation}

The Guassian basis function is adopted in this paper.

\begin{equation}
\phi_j(\textbf{x}) = \exp\left(-\frac{(x - \mu_j)^2}{2t^2}\right),
\end{equation}
where $\mu_j$ is the basis function location in input space, and $t$ is the spatial scale. The slip prediction is then compared with the set threshold for different levels of wheel slip, as discussed above. While slip risk is commonly defined in levels, the slip is terrain-dependent based on the estimated wheel traversing speed. The terrain-specific speed threshold is defined as $f(\phi)$, which is modified to $f(\phi - \sigma)$ in consideration of speed estimation accuracy as in (\ref{eq:speed_threshold}). 

\begin{equation}
\begin{aligned}
\label{eq:speed_threshold}
    s = f(\phi), \\
    s' = f(\phi - \sigma) = f(\phi'),
\end{aligned}
\end{equation}

\begin{equation}
    \text{MAE} : \sigma = \frac{1}{n} \sum_{i=1}^{n} |g_i - h_i|,
\end{equation}
where $g_i$ and $h_i$ are the estimated traversing speed and the ground truth value, respectively. Through the above techniques as adopted from \cite{endo2021terrain}, the terrain-dependent slip risk prediction threshold can be shifted from ($s, \phi$) to ($s', \phi'$)

\subsection{Data Generation} 
The Khalifa University Space Rover (KUSR), a differential drive grouser-wheeled rover \cite{yakubu4679693novel}, is used in a high-fidelity, physics-based simulator (Vortex Studio v22.8)\cite{cm2020vortex,10634320} for data collection. Egocentric RGB terrain images, right and left wheel angular speed, and slippage data are generated from the navigation of the rover on six heterogeneous terrains. These terrains are created by generating patterns using eight different types of soils, which include four lunar soil textures and four soils commonly found on Earth, as in Fig. \ref{fig:slip-speed_data}. Mechanical properties such as soil friction angle, cohesion, and stiffness modulus are assigned to the rover tire model. A total of 24 trajectories were generated with a $20m \times 20m$ terrain dimension. In total, there are 10080 input labels split into 8064/2016 training/testing sets, respectively.

\subsection{Model Architecture}

\begin{figure*}
    \centering
    \includegraphics[width=1\linewidth]{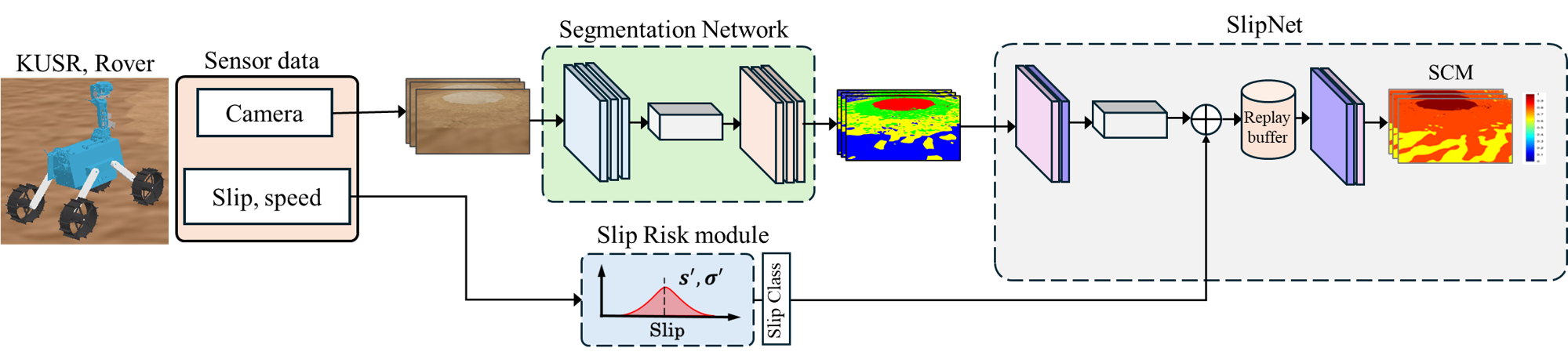}
    \caption{Our proposed scheme is constructed as a dual-network architecture including a segmentation network and the SlipNet}
    \label{fig:framework_overview}
\end{figure*}

The proposed SlipNet architecture consists of three major components as shown in Fig. \ref{fig:framework_overview}. \textit{Terrain segmentation module}, which converts a terrain image into a semantic segmentation map. Three semantic segmentation networks were explored in this study: U-Net, PSPNet, and DeepLab v3+. Due to its weak-supervision functionality, the DeepLab v3+ \cite{chen2018encoder} is adopted for segmentation. The \textit{Slip Risk Module} maps the segmentation mask with in-situ wheel slip mean and standard deviation values for each terrain type. The \textit{SlipNet} assigns slip risk classes to terrain segments and keeps track of all terrain classes and their corresponding slip risk estimates. In the SlipNet, the terrain segmented map is updated into a Slip Cost Map represented by jet colormap to indicate slippage risk zones.

\subsubsection{Terrain Semantic Segmentation}
\begin{figure}
    \centering
    \includegraphics[width=1.0\linewidth]{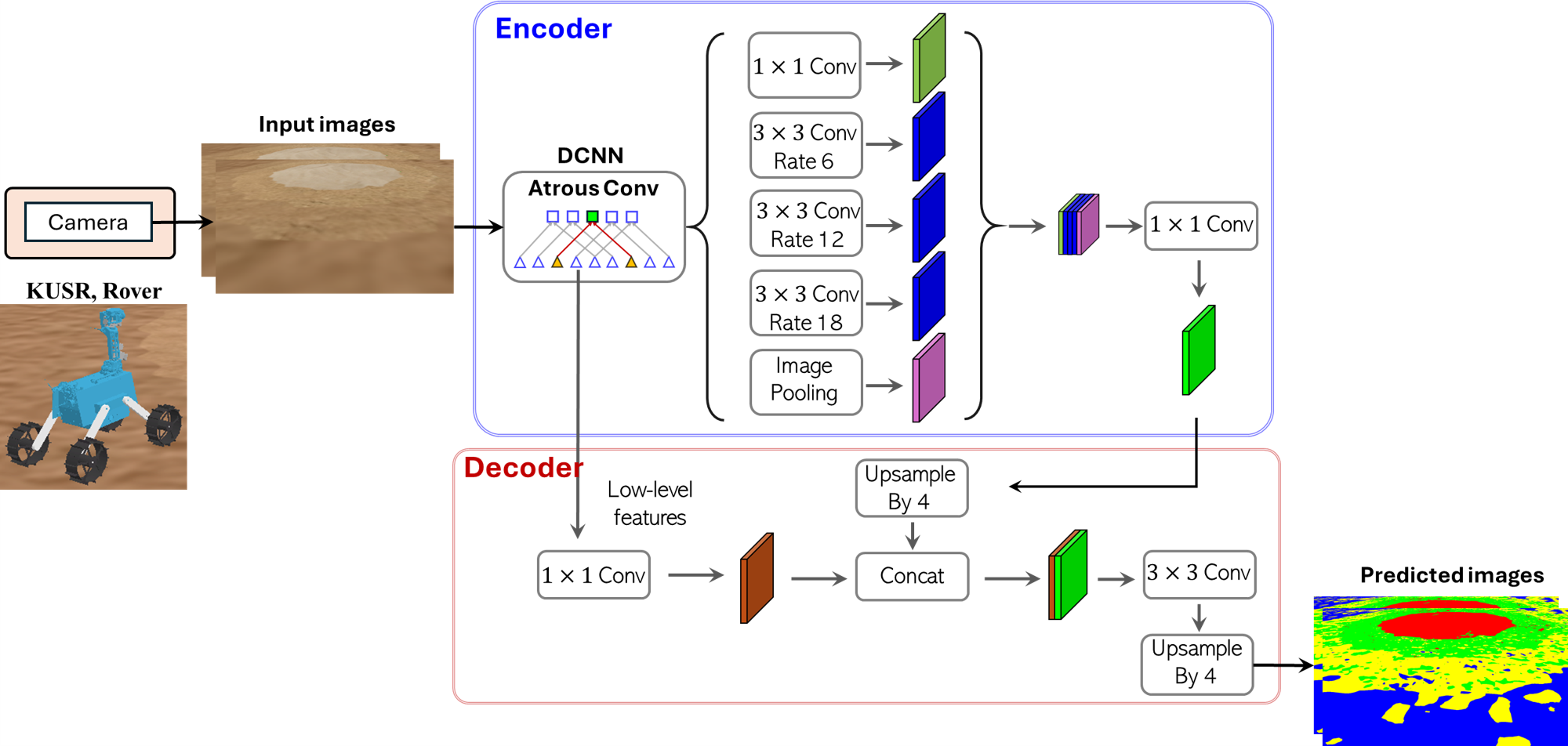}
    \caption{Overview of the encoder-decoder architecture of the DeepLab v3+}
    \label{fig:deeplabv3_overview}
\end{figure}

The terrain segmentation network utilizes the DeepLab implementation of fully convolutional neural network (FCNN) as described by \cite{chen2018encoder} and shown in Fig. \ref{fig:deeplabv3_overview}. The network's front end mirrors the VGG architecture but is adapted to incorporate "atrous" convolutions, also known as dilated convolutions. These specialized convolutions expand the receptive field of the filters without increasing the filter size.

To enhance the generalization of the segmentation models, only 70 \% of the terrain images were annotated with segmentation masks, creating a partial supervision setup. The network is trained using standard backpropagation and stochastic gradient descent, requiring approximately 6 hours on an Nvidia GTX 1050 GPU. The trained segmentation model processes an input terrain image within 125 milliseconds.

\subsubsection{SlipNet}
SlipNet is designed as a dynamic terrain analysis tool utilizing a Vision Transformer (ViT) encoder-decoder architecture to predict and adapt to varying slip conditions encountered by a rover during navigation. The input to the SlipNet consists of a segmentation map generated from the DeepLab v3+ model. The ViT encoder processes this map through self-attention mechanisms, capturing spatial hierarchies and inter-segment dependencies, with each segment treated as a token. A fusion layer then integrates real-time slip data from the rover's sensors. The ViT decoder uses these processed features to reconstruct an enhanced segmentation map that annotates each segment with predicted slip risks. SlipNet employs an experience replay buffer to store informative past experiences, using an attention mechanism to prioritize learning from segments that provide significant insight into terrain properties. 

SlipNet predicts both mean $m_i$ and variance $\sigma_i$  for each pixel $i$, using a negative Gaussian log-likelihood loss, averaged over valid pixels $i \in \mathcal{V}$ where ground truth $m_i^t$ is available \cite{wellhausen2019should}:

\begin{equation} 
\mathcal{L} = \sum_{i \in \mathcal{V}} \left( \frac{(m_i - m_i^t)^2}{2\sigma_i^2} + \log \sigma_i \right)
\end{equation}

\section{Results}
We evaluate our models on test datasets that consist of 2016 terrain images. The rover used for simulation experiments is KUSR, which is lightweight (about 10 kg) with four grouser wheels of about $0.1 m$ radius. We use a maximum angular speed range of $[-3.5, 3.5 ]$rad/s with the differential drive to guide the rover from the start position to the goal position for each path-following scenario. Each scenario takes around 30 seconds. To generate rich data across a variety of terrains, various heterogeneous terrains were created with different complexities using a mixture of eight different soil textures given in Fig. \ref{fig:slip-speed_data}.

\begin{table}[h]
\caption{Inference time and training time of our model (DeepLab v3+ + SlipNet) compared to the baseline methods executed on a workstation with NVidia GTX 1050 GPU. The training samples consist of 10,000 images resized to 256 X 256.}
\label{tab:method_comparison}
\centering
\resizebox{\columnwidth}{!}{%
\begin{tabular}{|c|c|}
\hline
\textbf{Method}          & \textbf{Inference Time (sec)} \\ \hline
U-Net + SlipNet  
& 0.110           \\ \hline
PSPNet + SlipNet         & 0.140             \\ \hline
DeepLab v3+ + SlipNet (Ours)  & 0.220           \\ \hline
TerrainNet \cite{meng2023terrainnet}                & 0.310 \\ \hline
\end{tabular}%
}
\end{table}

A sample of five heterogeneous terrain images was selected for both quantitative and qualitative performance comparison of our approach and the baselines. Adopting the pretrained segmentation models mainly for segmentation task along with SlipNet to predict segment slippage value was found to be more efficient in computational and time complexities. We investigated three pretrained segmentation models, which are U-Net, PSPNet, and DeepLab v3+. 
For a fair comparison with TerrainNet, the elevation loss component is replaced by slippage loss. TerrainNet takes the U-Net + SlipNet and takes the least inference time, about 0.11 seconds, for each terrain sample. DeepLab v3+ + SlipNet (Ours), despite having an inference time of about 0.22 seconds, has shown better prediction accuracy than the baselines in four out of five tested sample cases. Mean Absolute Error (MAE) is used for the quantitative comparison as shown in Table \ref{tab:quantitative_comparison}.

\begin{table*}[h]
\caption{Evaluation of slip hazard estimation performance for the different models studied and the baseline (TerrainNet) using MAE}
\label{tab:quantitative_comparison}
\centering
\begin{tabular}{|c|c|c|c|c|c|}
\hline
\textbf{Method}          & \textbf{Sample 1} & \textbf{Sample 2} & \textbf{Sample 3} & \textbf{Sample 4} & \textbf{Sample 5}  \\ \hline
U-Net + SlipNet              & 0.754            & 0.662            & \textbf{0.721 }             & 0.341             & 0.323                            \\ \hline
PSPNet + SlipNet         & 0.664             & 0.647             & 0.826              & 0.308              & 0.317                          \\ \hline

DeepLab v3+ + SlipNet  (Ours)   & \textbf{0.621}              & \textbf{0.617}             & 0.814              & \textbf{0.257}             & \textbf{0.315}                            \\ \hline

TerrainNet \cite{meng2023terrainnet}                & 0.796             & 0.771             &  0.919             & 0.380              &  0.379                          \\ \hline
\end{tabular}
\end{table*}

\begin{figure}
    \centering
    \includegraphics[width=1\linewidth]{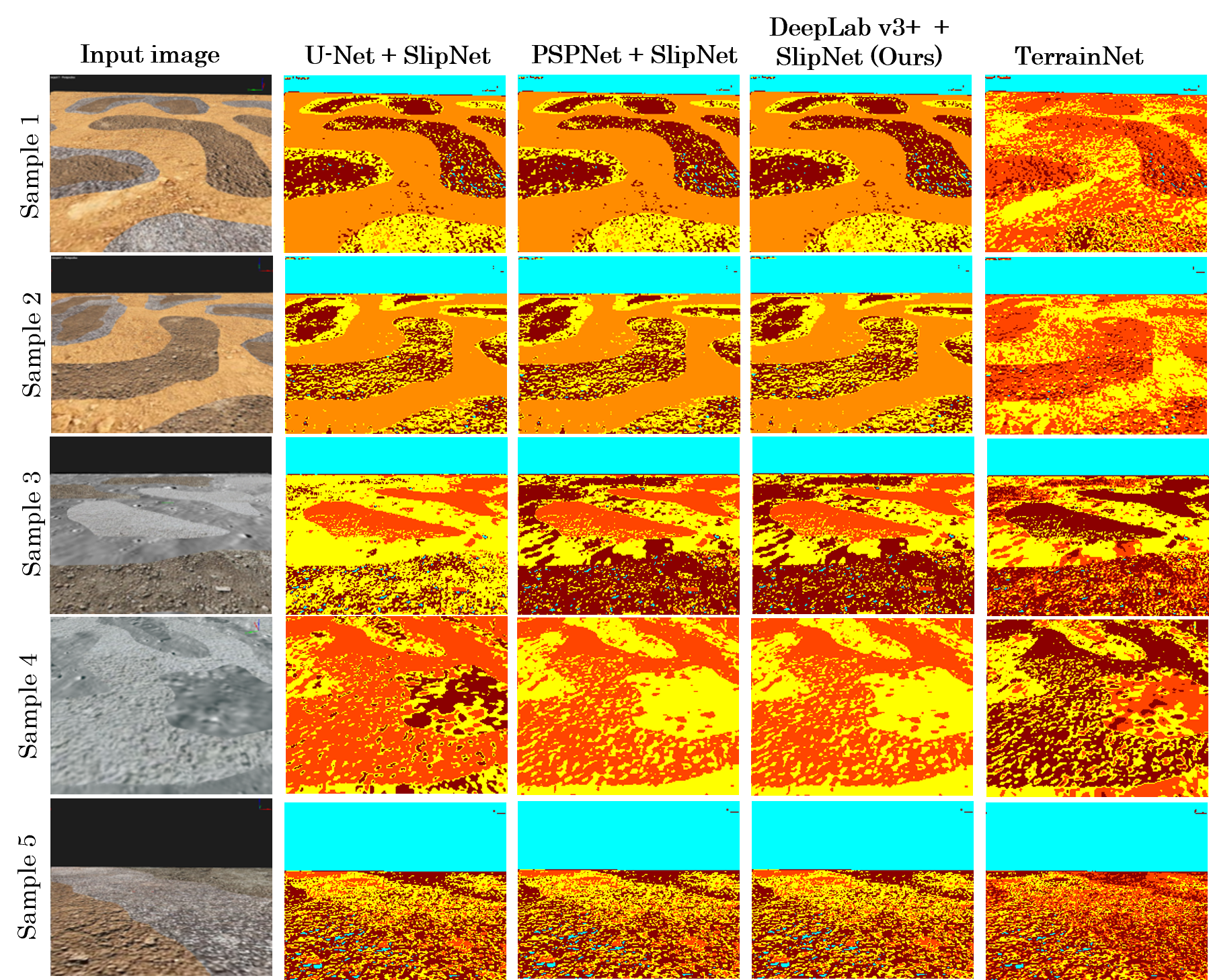}
    \caption{Qualitative comparison between different semantic segmentation networks combination with SlipNet and the baseline, in five different terrain samples. DeepLab v3+ is the best candidate in Slip cost map due to weak supervision which is necessary for segmenting terrain texture not included in the annotation mask}
    \label{fig:comparison_images}
\end{figure}

We evaluate the models qualitatively using the five heterogeneous image samples as shown in Fig. \ref{fig:comparison_images}. The slip risk module outputs five slip risk classes in the range of [0 to 1] slip ratio presented by colors in the Jet color map range. Untraversed terrains are initially assigned a moderate slip value range [0.2–0.4] represented by the cyan color ID. In all the five tested samples, TerrainNet performs the least in slip prediction performance for each terrain segment. U-Net + SlipNet performs best in sample 3. In general, slip prediction performance is better if there is a clear semantic distinction in the terrain image. This is expected since the performance of the semantic segmentation network is high with distinct image features. Unseen terrains with close visual outlook but different mechanical properties than the known terrains are hard to predict within the slip risk threshold. Sample 4, which mimics two types of heterogeneous lunar soil, was not encountered during training, but our model accurately predicts the slip values for both soils.

In our model (DeepLab v3+ + SlipNet),  resegmentation of unannotated terrain segment is handled through weak supervision mechanism of DeepLab v3+ network.

\begin{figure}
    \centering
    \includegraphics[width=1\linewidth]{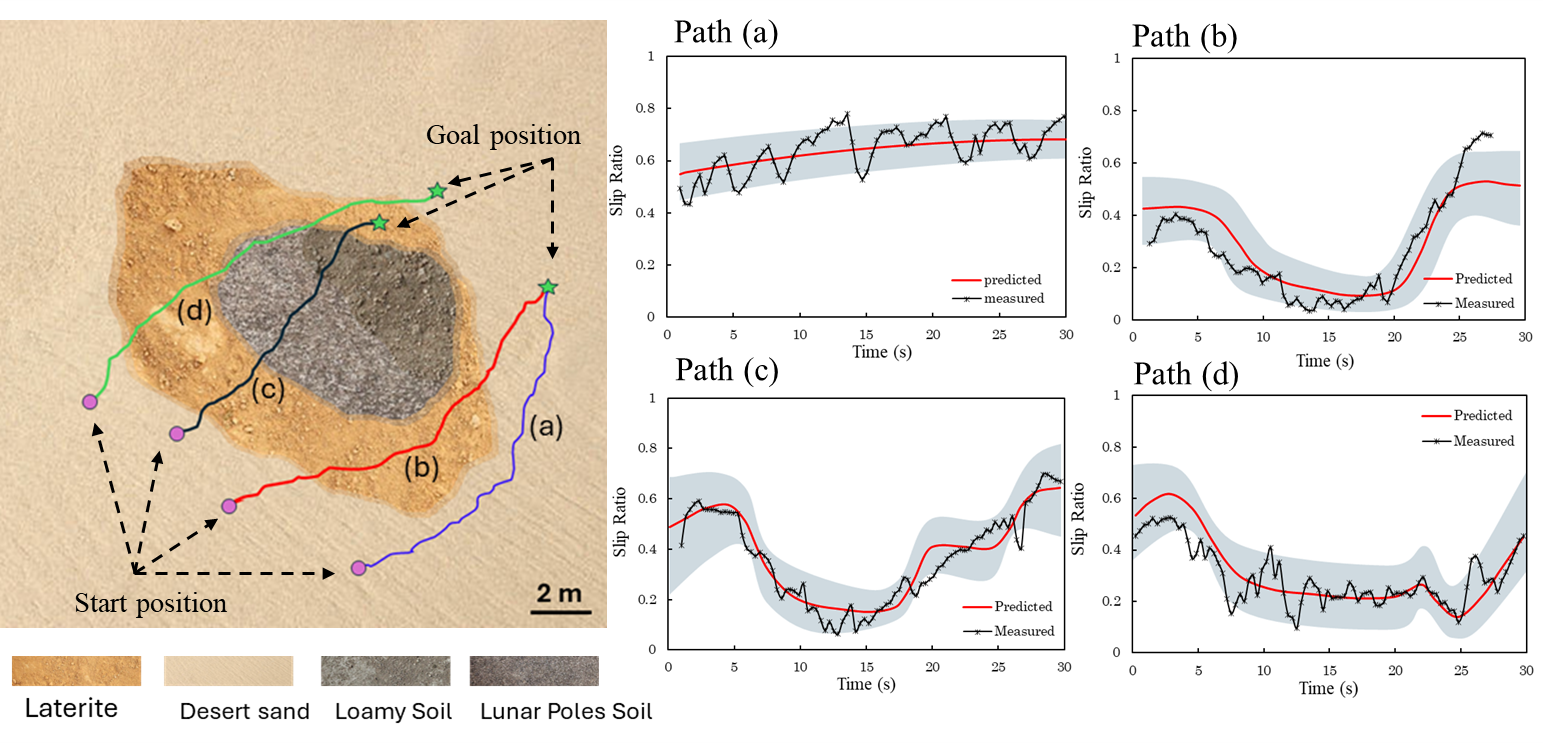}
    \caption{Slip predictions using our model (DeepLab v3+ + SlipNet) under four different paths in a heterogeneous terrain environment. The red curve represents the predictive mean slip value. The blue-shaded area represents the standard deviations around the mean. The black curve represented the actual measured slippage.}
    \label{fig:four_path_slip}
\end{figure}

The slip prediction performance of our model is tested in a new heterogeneous terrain. Four paths were followed under a rover speed of about $[0.3 - 0.5]$ m/s traversing different soils with different mechanical properties. As shown in Fig. \ref{fig:four_path_slip}, the predicted mean slip
is represented by a red curve, and the standard deviation values are represented by a blue-shaded area around the mean. The actual mean slip value measured from the rover is plotted in the black curve. The slip mean value is computed from the average of right and left wheel slip ratios.
 During motion on desert sand-like terrain in path (a), our model predicts a high slip ratio of around 0.6. The measured average slip ratio value fluctuates due to variations in speeds from the right and left wheels in achieving differential steering drive on the soft desert sand-like terrain. 
 In paths (b), (c), and (d), multiple terrain textures were traversed and the measured slip value was predicted well within the prediction standard deviation. However, relatively high deviations experienced in some steps within the test scenarios are due to sudden changes in speed which were not handled by our current model. 

 All test scenarios for path-following experiments were conducted with a simple PD controller that does not take into consideration terrain slip condition. This model is suitable for generating a slip cost map for incorporation with an autonomous navigation algorithm. 
 
\section{Conclusions}

In this study, we presented SlipNet, a model for predicting wheel slip in deformable terrain environments without the need for prior terrain classification. Our method employs adaptive reclassification, allowing for dynamic segmentation and classification of previously unseen terrains during deployment. Through extensive simulations using datasets generated by our high-fidelity synthetic data framework based on Vortex Studio, we demonstrated that SlipNet significantly outperforms the state-of-the-art TerrainNet method across all five test scenarios. These results underscore the effectiveness of SlipNet in handling varying soil properties and enhancing rover navigation capabilities in uncertain terrains. Future research will focus on integrating the slip cost map generated by SlipNet into learning-based autonomous navigation systems to enable real-time decision-making in complex environments.\\

\section{Acknowledgements.} 
We extend our sincere gratitude to the entire team at the Khalifa University Center for Autonomous Robotic Systems (KUCARS) and Advanced Research and Innovation Center (ARIC) for their invaluable support and collaboration throughout this research. Our gratitude also extends to Khalifa University of Science and Technology for providing the essential funding and resources that were crucial for the successful completion of this work.

\bibliography{paper}

\begin{thebibliography}{10}
\providecommand{\url}[1]{#1}
\csname url@samestyle\endcsname
\providecommand{\newblock}{\relax}
\providecommand{\bibinfo}[2]{#2}
\providecommand{\BIBentrySTDinterwordspacing}{\spaceskip=0pt\relax}
\providecommand{\BIBentryALTinterwordstretchfactor}{4}
\providecommand{\BIBentryALTinterwordspacing}{\spaceskip=\fontdimen2\font plus
\BIBentryALTinterwordstretchfactor\fontdimen3\font minus \fontdimen4\font\relax}
\providecommand{\BIBforeignlanguage}[2]{{%
\expandafter\ifx\csname l@#1\endcsname\relax
\typeout{** WARNING: IEEEtran.bst: No hyphenation pattern has been}%
\typeout{** loaded for the language `#1'. Using the pattern for}%
\typeout{** the default language instead.}%
\else
\language=\csname l@#1\endcsname
\fi
#2}}
\providecommand{\BIBdecl}{\relax}
\BIBdecl

\bibitem{quadrelli2015guidance}
M.~B. Quadrelli, L.~J. Wood, J.~E. Riedel, M.~C. McHenry, M.~Aung, L.~A. Cangahuala, R.~A. Volpe, P.~M. Beauchamp, and J.~A. Cutts, ``Guidance, navigation, and control technology assessment for future planetary science missions,'' \emph{Journal of Guidance, Control, and Dynamics}, vol.~38, no.~7, pp. 1165--1186, 2015.

\bibitem{toupet2020driving}
O.~Toupet, D.~Levine, N.~Patel, J.~Biesiadecki, M.~Maimone, and A.~Rankin, ``Driving curiosity: Mars rover mobility trends during the first 7 years,'' 2020.

\bibitem{arvidson2017mars}
R.~E. Arvidson, K.~D. Iagnemma, M.~Maimone, A.~A. Fraeman, F.~Zhou, M.~C. Heverly, P.~Bellutta, D.~Rubin, N.~T. Stein, J.~P. Grotzinger \emph{et~al.}, ``Mars science laboratory curiosity rover megaripple crossings up to sol 710 in gale crater,'' \emph{Journal of Field Robotics}, vol.~34, no.~3, pp. 495--518, 2017.

\bibitem{gonzalez2018slippage}
R.~Gonzalez and K.~Iagnemma, ``Slippage estimation and compensation for planetary exploration rovers. state-of-the-art and future challenges,'' \emph{Journal of Field Robotics}, vol.~35, no.~4, pp. 564--577, 2018.

\bibitem{hutangkabodee2006performance}
S.~Hutangkabodee, Y.~H. Zweiri, L.~D. Seneviratne, and K.~Althoefer, ``Performance prediction of a wheeled vehicle on unknown terrain using identified soil parameters,'' in \emph{Proceedings 2006 IEEE International Conference on Robotics and Automation, 2006. ICRA 2006.}\hskip 1em plus 0.5em minus 0.4em\relax IEEE, 2006, pp. 3356--3361.

\bibitem{10406669}
A.~Abubakar, R.~Alhammadi, Y.~Zweiri, and L.~Seneviratne, ``Advance simulation method for wheel-terrain interactions of space rovers: A case study on the uae rashid rover*,'' in \emph{2023 21st International Conference on Advanced Robotics (ICAR)}.\hskip 1em plus 0.5em minus 0.4em\relax IEEE, 2023, pp. 526--532.

\bibitem{song2009slip}
Z.~Song, L.~D. Seneviratne, K.~Althoefer, X.~Song, and Y.~H. Zweiri, ``Slip parameter estimation of a single wheel using a non-linear observer,'' \emph{Robotica}, vol.~27, no.~6, pp. 801--811, 2009.

\bibitem{guan2022ga}
T.~Guan, D.~Kothandaraman, R.~Chandra, A.~J. Sathyamoorthy, K.~Weerakoon, and D.~Manocha, ``Ga-nav: Efficient terrain segmentation for robot navigation in unstructured outdoor environments,'' \emph{IEEE Robotics and Automation Letters}, vol.~7, no.~3, pp. 8138--8145, 2022.

\bibitem{guan2021tns}
T.~Guan, Z.~He, R.~Song, D.~Manocha, and L.~Zhang, ``Tns: Terrain traversability mapping and navigation system for autonomous excavators,'' \emph{arXiv preprint arXiv:2109.06250}, 2021.

\bibitem{maturana2018real}
D.~Maturana, P.-W. Chou, M.~Uenoyama, and S.~Scherer, ``Real-time semantic mapping for autonomous off-road navigation,'' in \emph{Field and Service Robotics: Results of the 11th International Conference}.\hskip 1em plus 0.5em minus 0.4em\relax Springer, 2018, pp. 335--350.

\bibitem{wellhausen2019should}
L.~Wellhausen, A.~Dosovitskiy, R.~Ranftl, K.~Walas, C.~Cadena, and M.~Hutter, ``Where should i walk? predicting terrain properties from images via self-supervised learning,'' \emph{IEEE Robotics and Automation Letters}, vol.~4, no.~2, pp. 1509--1516, 2019.

\bibitem{rothrock2016spoc}
B.~Rothrock, R.~Kennedy, C.~Cunningham, J.~Papon, M.~Heverly, and M.~Ono, ``Spoc: Deep learning-based terrain classification for mars rover missions,'' in \emph{AIAA SPACE 2016}, 2016, p. 5539.

\bibitem{meng2023terrainnet}
X.~Meng, N.~Hatch, A.~Lambert, A.~Li, N.~Wagener, M.~Schmittle, J.~Lee, W.~Yuan, Z.~Chen, S.~Deng \emph{et~al.}, ``Terrainnet: Visual modeling of complex terrain for high-speed, off-road navigation,'' \emph{arXiv preprint arXiv:2303.15771}, 2023.

\bibitem{heverly2013traverse}
M.~Heverly, J.~Matthews, J.~Lin, D.~Fuller, M.~Maimone, J.~Biesiadecki, and J.~Leichty, ``Traverse performance characterization for the mars science laboratory rover,'' \emph{Journal of Field Robotics}, vol.~30, no.~6, pp. 835--846, 2013.

\bibitem{goldberg2002stereo}
S.~B. Goldberg, M.~W. Maimone, and L.~Matthies, ``Stereo vision and rover navigation software for planetary exploration,'' in \emph{Proceedings, IEEE aerospace conference}, vol.~5.\hskip 1em plus 0.5em minus 0.4em\relax IEEE, 2002, pp. 5--5.

\bibitem{maimone2007two}
M.~Maimone, Y.~Cheng, and L.~Matthies, ``Two years of visual odometry on the mars exploration rovers,'' \emph{Journal of Field Robotics}, vol.~24, no.~3, pp. 169--186, 2007.

\bibitem{estlin2012aegis}
T.~A. Estlin, B.~J. Bornstein, D.~M. Gaines, R.~C. Anderson, D.~R. Thompson, M.~Burl, R.~Castano, and M.~Judd, ``Aegis automated science targeting for the mer opportunity rover,'' \emph{ACM Transactions on Intelligent Systems and Technology (TIST)}, vol.~3, no.~3, pp. 1--19, 2012.

\bibitem{wagstaff2013smart}
K.~Wagstaff, D.~Thompson, W.~Abbey, A.~Allwood, D.~Bekker, N.~Cabrol, T.~Fuchs, and K.~Ortega, ``Smart, texture-sensitive instrument classification for in situ rock and layer analysis,'' \emph{Geophysical Research Letters}, vol.~40, no.~16, pp. 4188--4193, 2013.

\bibitem{gupta2021ocrnet}
V.~Gupta, A.~Gupta, N.~Arora, and J.~Garg, ``Ocrnet-light-weighted and efficient neural network for optical character recognition,'' in \emph{2021 IEEE Bombay Section Signature Conference (IBSSC)}.\hskip 1em plus 0.5em minus 0.4em\relax IEEE, 2021, pp. 1--4.

\bibitem{zhao2017pyramid}
H.~Zhao, J.~Shi, X.~Qi, X.~Wang, and J.~Jia, ``Pyramid scene parsing network,'' in \emph{Proceedings of the IEEE conference on computer vision and pattern recognition}, 2017, pp. 2881--2890.

\bibitem{dai2022segmarsvit}
Y.~Dai, T.~Zheng, C.~Xue, and L.~Zhou, ``Segmarsvit: Lightweight mars terrain segmentation network for autonomous driving in planetary exploration,'' \emph{Remote Sensing}, vol.~14, no.~24, p. 6297, 2022.

\bibitem{viswanath2021offseg}
K.~Viswanath, K.~Singh, P.~Jiang, P.~Sujit, and S.~Saripalli, ``Offseg: A semantic segmentation framework for off-road driving,'' in \emph{2021 IEEE 17th international conference on automation science and engineering (CASE)}.\hskip 1em plus 0.5em minus 0.4em\relax IEEE, 2021, pp. 354--359.

\bibitem{roth2023viplanner}
P.~Roth, J.~Nubert, F.~Yang, M.~Mittal, and M.~Hutter, ``Viplanner: Visual semantic imperative learning for local navigation,'' \emph{arXiv preprint arXiv:2310.00982}, 2023.

\bibitem{shaban2022semantic}
A.~Shaban, X.~Meng, J.~Lee, B.~Boots, and D.~Fox, ``Semantic terrain classification for off-road autonomous driving,'' in \emph{Conference on Robot Learning}.\hskip 1em plus 0.5em minus 0.4em\relax PMLR, 2022, pp. 619--629.

\bibitem{wigness2019rugd}
M.~Wigness, S.~Eum, J.~G. Rogers, D.~Han, and H.~Kwon, ``A rugd dataset for autonomous navigation and visual perception in unstructured outdoor environments,'' in \emph{2019 IEEE/RSJ International Conference on Intelligent Robots and Systems (IROS)}.\hskip 1em plus 0.5em minus 0.4em\relax IEEE, 2019, pp. 5000--5007.

\bibitem{jiang2021rellis}
P.~Jiang, P.~Osteen, M.~Wigness, and S.~Saripalli, ``Rellis-3d dataset: Data, benchmarks and analysis,'' in \emph{2021 IEEE International Conference on robotics and automation (ICRA)}.\hskip 1em plus 0.5em minus 0.4em\relax IEEE, 2021, pp. 1110--1116.

\bibitem{valada2017deep}
A.~Valada, G.~L. Oliveira, T.~Brox, and W.~Burgard, ``Deep multispectral semantic scene understanding of forested environments using multimodal fusion,'' in \emph{2016 International Symposium on Experimental Robotics}.\hskip 1em plus 0.5em minus 0.4em\relax Springer, 2017, pp. 465--477.

\bibitem{bradley2015scene}
D.~M. Bradley, J.~K. Chang, D.~Silver, M.~Powers, H.~Herman, P.~Rander, and A.~Stentz, ``Scene understanding for a high-mobility walking robot,'' in \emph{2015 IEEE/RSJ International Conference on Intelligent Robots and Systems (IROS)}.\hskip 1em plus 0.5em minus 0.4em\relax IEEE, 2015, pp. 1144--1151.

\bibitem{schilling2017geometric}
F.~Schilling, X.~Chen, J.~Folkesson, and P.~Jensfelt, ``Geometric and visual terrain classification for autonomous mobile navigation,'' in \emph{2017 IEEE/RSJ International Conference on Intelligent Robots and Systems (IROS)}.\hskip 1em plus 0.5em minus 0.4em\relax IEEE, 2017, pp. 2678--2684.

\bibitem{ono2015risk}
M.~Ono, T.~J. Fuchs, A.~Steffy, M.~Maimone, and J.~Yen, ``Risk-aware planetary rover operation: Autonomous terrain classification and path planning,'' in \emph{2015 IEEE aerospace conference}.\hskip 1em plus 0.5em minus 0.4em\relax IEEE, 2015, pp. 1--10.

\bibitem{swan2021ai4mars}
R.~M. Swan, D.~Atha, H.~A. Leopold, M.~Gildner, S.~Oij, C.~Chiu, and M.~Ono, ``Ai4mars: A dataset for terrain-aware autonomous driving on mars,'' in \emph{Proceedings of the IEEE/CVF conference on computer vision and pattern recognition}, 2021, pp. 1982--1991.

\bibitem{zhang2022s}
J.~Zhang, L.~Lin, Z.~Fan, W.~Wang, and J.~Liu, ``Mars: Self-supervised and semi-supervised learning for mars segmentation,'' \emph{arXiv preprint arXiv:2207.01200}, 2022.

\bibitem{endo2023risk}
M.~Endo, T.~Taniai, R.~Yonetani, and G.~Ishigami, ``Risk-aware path planning via probabilistic fusion of traversability prediction for planetary rovers on heterogeneous terrains,'' in \emph{2023 IEEE international conference on robotics and automation (ICRA)}.\hskip 1em plus 0.5em minus 0.4em\relax IEEE, 2023, pp. 11\,852--11\,858.

\bibitem{cai2023probabilistic}
X.~Cai, M.~Everett, L.~Sharma, P.~R. Osteen, and J.~P. How, ``Probabilistic traversability model for risk-aware motion planning in off-road environments,'' in \emph{2023 IEEE/RSJ International Conference on Intelligent Robots and Systems (IROS)}.\hskip 1em plus 0.5em minus 0.4em\relax IEEE, 2023, pp. 11\,297--11\,304.

\bibitem{otsu2016autonomous}
K.~Otsu, M.~Ono, T.~J. Fuchs, I.~Baldwin, and T.~Kubota, ``Autonomous terrain classification with co-and self-training approach,'' \emph{IEEE Robotics and Automation Letters}, vol.~1, no.~2, pp. 814--819, 2016.

\bibitem{castro2023does}
M.~G. Castro, S.~Triest, W.~Wang, J.~M. Gregory, F.~Sanchez, J.~G. Rogers, and S.~Scherer, ``How does it feel? self-supervised costmap learning for off-road vehicle traversability,'' in \emph{2023 IEEE International Conference on Robotics and Automation (ICRA)}.\hskip 1em plus 0.5em minus 0.4em\relax IEEE, 2023, pp. 931--938.

\bibitem{richter2017safe}
C.~Richter and N.~Roy, ``Safe visual navigation via deep learning and novelty detection,'' 2017.

\bibitem{frey2023fast}
J.~Frey, M.~Mattamala, N.~Chebrolu, C.~Cadena, M.~Fallon, and M.~Hutter, ``Fast traversability estimation for wild visual navigation,'' \emph{arXiv preprint arXiv:2305.08510}, 2023.

\bibitem{schmid2022self}
R.~Schmid, D.~Atha, F.~Sch{\"o}ller, S.~Dey, S.~Fakoorian, K.~Otsu, B.~Ridge, M.~Bjelonic, L.~Wellhausen, M.~Hutter \emph{et~al.}, ``Self-supervised traversability prediction by learning to reconstruct safe terrain,'' in \emph{2022 IEEE/RSJ International Conference on Intelligent Robots and Systems (IROS)}.\hskip 1em plus 0.5em minus 0.4em\relax IEEE, 2022, pp. 12\,419--12\,425.

\bibitem{xue2023contrastive}
H.~Xue, X.~Hu, R.~Xie, H.~Fu, L.~Xiao, Y.~Nie, and B.~Dai, ``Contrastive label disambiguation for self-supervised terrain traversability learning in off-road environments,'' \emph{arXiv preprint arXiv:2307.02871}, 2023.

\bibitem{gasparino2022wayfast}
M.~V. Gasparino, A.~N. Sivakumar, Y.~Liu, A.~E. Velasquez, V.~A. Higuti, J.~Rogers, H.~Tran, and G.~Chowdhary, ``Wayfast: Navigation with predictive traversability in the field,'' \emph{IEEE Robotics and Automation Letters}, vol.~7, no.~4, pp. 10\,651--10\,658, 2022.

\bibitem{alhammadi2024event}
R.~Alhammadi, Y.~Zweiri, A.~Abubakar, M.~Yakubu, L.~Abuassi, and L.~Seneviratne, ``Event-based slip estimation framework for space rovers traversing soft terrains,'' \emph{IEEE Access}, 2024.

\bibitem{10634320}
A.~Abubakar, Y.~Zweiri, R.~Alhammadi, M.~B. Mohiuddin, M.~Yakubu, and L.~Seneviratne, ``Predictor-based control for delay compensation in bilateral teleoperation of wheeled rovers on soft terrains,'' \emph{IEEE Access}, vol.~12, pp. 111\,593--111\,610, 2024.

\bibitem{endo2021terrain}
M.~Endo, S.~Endo, K.~Nagaoka, and K.~Yoshida, ``Terrain-dependent slip risk prediction for planetary exploration rovers,'' \emph{Robotica}, vol.~39, no.~10, pp. 1883--1896, 2021.

\bibitem{yakubu4679693novel}
M.~Yakubu, Y.~Zweiri, L.~AbuAssi, R.~Azzam, A.~Busoud, and L.~Seneviratne, ``A novel mobility concept for terrestrial wheel-legged lunar rover,'' \emph{Available at SSRN 4679693}.

\bibitem{cm2020vortex}
CM-Labs, ``Vortex studio,'' \emph{CM Labs}, 2020.

\bibitem{chen2018encoder}
L.-C. Chen, Y.~Zhu, G.~Papandreou, F.~Schroff, and H.~Adam, ``Encoder-decoder with atrous separable convolution for semantic image segmentation,'' in \emph{Proceedings of the European conference on computer vision (ECCV)}, 2018, pp. 801--818.

\end{thebibliography}

\end{document}